\documentclass[a4paper,conference]{IEEEtran}

\usepackage{amssymb}
\usepackage{amstext}
\usepackage{amsmath}
\usepackage{amsthm}
\usepackage{multicol, multirow}
\usepackage[pdftex]{graphicx}
\usepackage{cite}
\usepackage{dsfont}
\usepackage{tikz}
\usepackage{pifont}
\usepackage{overpic}
\usepackage{tikz}
\usetikzlibrary{arrows.meta, positioning}
\usetikzlibrary{arrows, shapes, backgrounds, fit} 
\usetikzlibrary{shadows, positioning}
\usepackage{circuitikz}
\usepackage{tikzscale}
\usepackage{pstricks}
\usepackage{pgfplots}
\newlength\figureheight 
\newlength\figurewidth
\usepackage{soul}
\usepackage{tabu}
\usepackage{url}
\usepackage{booktabs}

\usepackage{microtype}

\newcommand{\cmark}{\ding{51}}

\DeclareMathOperator*{\argmax}{argmax}

\begin{document}

\title{Distinctive 3D local deep descriptors}

\author{
\IEEEauthorblockN{Fabio Poiesi and Davide Boscaini}
\IEEEauthorblockA{Technologies of Vision, Fondazione Bruno Kessler, Trento, Italy\\
\{poiesi, dboscaini\}@fbk.eu
}}

\maketitle

\begin{abstract}
We present a simple but yet effective method for learning distinctive 3D local deep descriptors (DIPs) that can be used to register point clouds without requiring an initial alignment.
Point cloud patches are extracted, canonicalised with respect to their estimated local reference frame and encoded into rotation-invariant compact descriptors by a PointNet-based deep neural network.
DIPs can effectively generalise across different sensor modalities because they are learnt end-to-end from locally and randomly sampled points.
Because DIPs encode only local geometric information, they are robust to clutter, occlusions and missing regions.
We evaluate and compare DIPs against alternative hand-crafted and deep descriptors on several indoor and outdoor datasets consisting of point clouds reconstructed using different sensors.
Results show that DIPs (i) achieve comparable results to the state-of-the-art on RGB-D indoor scenes (3DMatch dataset), (ii) outperform state-of-the-art by a large margin on laser-scanner outdoor scenes (ETH dataset), and (iii) generalise to indoor scenes reconstructed with the Visual-SLAM system of Android ARCore.
Source code: \url{https://github.com/fabiopoiesi/dip}.
\end{abstract}

\IEEEpeerreviewmaketitle

\section{Introduction}

\begin{figure}[t]
  \centering
  \includegraphics[width=1\columnwidth]{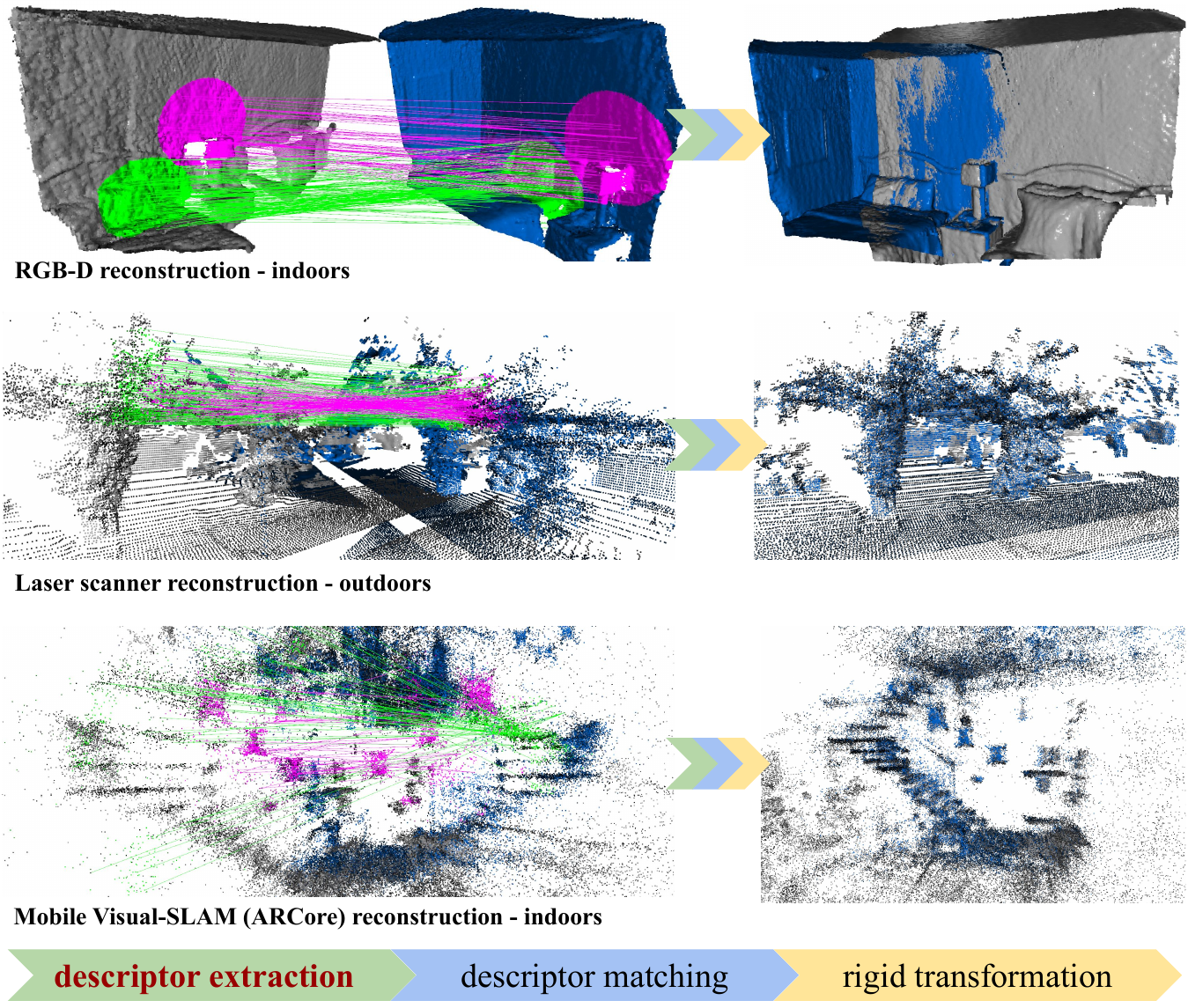}
  \vspace{-6mm}
  \caption{
  Point cloud registration typically involves three steps: description extraction, descriptor matching and rigid transformation. 
  We focus on the first step.
  DIPs can be used to register point clouds reconstructed with different sensor modalities in different environments. 
  DIPs are local, rotation-invariant and compact descriptors that are learnt from examples using a PointNet-based deep neural network \cite{Qi2017a}. 
  The correspondences between points of corresponding patches (magenta and green) are implicitly learnt by the network. 
  The key aspect of DIPs is their generalisation ability.}
  \label{fig:teaser}
\end{figure}

Encoding local 3D geometric information (e.g.~coordinates, normals) into compact descriptors is key for shape retrieval \cite{Ovsjanikov2010}, face recognition \cite{Lei2016}, object recognition \cite{Johnson1999} and rigid (six degrees-of-freedom) registration \cite{Zeng2017}.
Learning such encoding from examples using deep neural networks has outperformed hand-crafted methods \cite{Zeng2017,Yang2017,Deng2018cvpr,Deng2018eccv,Deng2019,Zhao2019,Gojcic2019,Choy2019,Bai2020}.
These approaches have been designed to encode geometric information either from meshes \cite{Boscaini2016,Monti2017} or from point clouds \cite{Gojcic2019,Choy2019,Bai2020}.
Our method belongs to the latter category and can be used to rigidly register point clouds without requiring an initial alignment (Fig.~\ref{fig:teaser}).

Existing solutions to compute compact 3D descriptors can be categorised into one-stage \cite{Rusu2009,Zeng2017,Deng2018cvpr,Choy2019,Bai2020} and two-stage \cite{Johnson1999,Salti2014,Gojcic2019} methods.
Although both categories share the objective of making descriptors invariant to point-cloud rigid transformations, one-stage methods encode the local geometric information of a patch (collection of locally sampled points) using the points of the patch directly.
Differently, two-stage methods firstly estimate a local reference frame (LRF) from the points within the patch to rigidly transform the patch to a canonical frame, then they encode the information of the canonicalised points into a compact descriptor.
Given two corresponding patches of two non-aligned point clouds, if we canonicalise them through their respective LRFs we should obtain two identical overlapping patches.
Hence, the encoding method should be simpler to design than that of one-stage methods.
However, noise, occlusions and point clouds reconstructed with different sensors make the LRF estimation challenging \cite{Melzi2019,Zhu2020}.
Modelling descriptors to be robust to different sensors (e.g.~RGB-D, laser scanner) and to different environments (e.g.~indoor, outdoor) is also a challenge \cite{Gojcic2019,Bai2020}.
One-stage learning-based methods can achieve rotation invariance by encoding the local geometric information with a set of geometric relationships, such as points, normals and point pair features \cite{Deng2018cvpr}, and then by learning descriptors via a PointNet-based deep network in order to achieve permutation invariance with respect to the set of the input points \cite{Qi2017a}.
Alternatively, 3D convolutional neural networks (ConvNets) can be used locally, to process patches around interest points \cite{Zeng2017}, or globally, to process whole point clouds \cite{Choy2019}.
In two-step methods, LRFs can be computed with hand-crafted \cite{Yang2016} or learning-based \cite{Melzi2019} methods.
After LRF canonicalisation, points can be transformed into a voxel grid, where each voxel encodes the density of the points within \cite{Gojcic2019}.
Then, descriptors can be encoded from this voxel representations using a 3D ConvNet learnt with a Siamese approach \cite{Tian2017}.

In this paper we present a novel two-stage method where compact descriptors are learnt end-to-end from canonicalised patches.
To mitigate the problem of incorrectly estimated LRFs, we learn an affine transformation that refines the canonicalisation operation by minimising the Euclidean distance between points through the Chamfer loss \cite{Zhao2019}.
Similarly to \cite{Deng2018cvpr}, we learn descriptors with a PointNet-based deep neural network through a Siamese approach, but differently from \cite{Deng2018cvpr} 
(i) we use LRFs to canonicalise patches, 
(ii) our descriptors encode local information only, thus promoting robustness to clutter, occlusions, and missing regions, and
(iii) we use a hardest contrastive loss to mine for quadruplets \cite{Choy2019}, thus improving metric learning.
Differently from \cite{Deng2018cvpr} and \cite{Gojcic2019}, points are consumed directly by our network without adding augmented hand-crafted features or performing prior voxelisations.
We train our network using the 3DMatch dataset that consists of indoor scenes reconstructed with RGB-D sensors \cite{Zeng2017}.
We achieve state-of-the-art results on the 3DMatch test set and on its augmented version, namely 3DMatchRotated \cite{Deng2018eccv}, employed to assess descriptor rotation invariance.
We significantly outperform existing approaches in terms of generalisation ability to different sensor modalities (RGB-D $\!\rightarrow\!$ laser scanner) and to different environments (indoor bedrooms $\!\rightarrow\!$ outdoor forest) using the ETH dataset \cite{Pomerleau2012}.
Moreover, we validate DIP generalisation ability to another sensor modality (RGB-D $\!\rightarrow\!$ smartphone) by capturing three overlapping indoor point clouds with the Visual-SLAM system \cite{Mur-Artal2015} of an ARCore-based App \cite{arcore} we have developed to reconstruct the environment.
Notably, DIPs can successfully and robustly be used also to align these point clouds.
The source code and the reconstruction App are publicly available.

\section{Our approach}

Given a point cloud $\mathcal{P} \subset \mathbb{R}^3$, we define a local \emph{patch} $\mathcal{X}  = \{\mathbf{x}\} \subset \mathcal{P}$ as an unordered set of 3D points $\mathbf{x}$ with cardinality $\lvert \mathcal{X} \rvert = n$.
We design a deep neural network $\Phi_{\boldsymbol{\Theta}}$ that generates DIPs such that $\mathbf{f} = \Phi_{\boldsymbol{\Theta}}(\mathcal{X})$, where $\mathbf{f} \subset \mathbb{R}^d$ and $\boldsymbol{\Theta}$ is the set of learnable parameters.
Without loss of generality we use the 3D coordinates of the points as input to $\Phi_{\boldsymbol{\Theta}}$, i.e.~$\mathbf{x} = (x, y, z)$.

\subsection{Network architecture}

\begin{figure}[t]
  \centering
  \resizebox{\linewidth}{!}{\begin{tikzpicture}[
	>=stealth',
	fixed/.style={
		shape=rectangle, rounded corners,
		fill=black!5,
		draw=black!20, 
		text=black,
		font=\fontsize{12}{12}\selectfont,
		align=center
	},
	input/.style={
		shape=rectangle, rounded corners,
		fill=black!25!green!10,
		draw=black!25!green!40, 
		text=black,
		font=\fontsize{12}{12}\selectfont,
		align=center
	},
	param/.style={
		shape=rectangle, 
		rounded corners,
		fill=blue!15,
		draw=blue!40, 
		text=black,
		font=\fontsize{12}{12}\selectfont,
		align=center
	},
	output/.style={
		shape=rectangle, 
		rounded corners,
		fill=black!20!red!10,
		draw=black!20!red!40, 
		text=black,
		font=\fontsize{12}{12}\selectfont,
		align=center
	},
	feat/.style={
		shape=rectangle, 
		rounded corners,
		fill=black!15!yellow!15,
		draw=black!15!orange!30, 
		text=black,
		font=\fontsize{12}{12}\selectfont,
		align=center
	},
	oper/.style={
		shape=rectangle,
		rounded corners,
		fill=white!15, 
		draw=black!40, 
		text=black,
		font=\fontsize{12}{12}\selectfont,
		align=center, 
	},
	arrow/.style={
		-latex,
		color=black,
		draw=black,
		font=\fontsize{8}{8}\selectfont
	}
]

\coordinate (A0);
\node[fixed, right=0mm of A0] (B0) {\rotatebox{90}{$\mathcal{X}, (n, 3)$}};
\coordinate[right=2mm of B0] (C0);
\coordinate[above=15mm of C0] (C1);
\node[param, right=2.5mm of C1] (C2) {\rotatebox{90}{$\text{TNet}$}};
\node[fixed, right=2.5mm of C2] (C3) {\rotatebox{90}{$\mathbf{A}, (3, 3)$}};
\coordinate[right=2.5mm of C3] (C4);
\node[oper, right=16.75mm of C0] (D0) {\rotatebox{90}{$\text{matmul}$}};
\node[fixed, right=2.5mm of D0] (D1) {\rotatebox{90}{$\hat{\mathcal{X}}, (n, 3)$}};
\coordinate[right=2.5mm of D1] (D2);
\node[param, right=2.5mm of D2] (D2a) {$\text{MLP}_1$};
\coordinate[right=2.5mm of D2a] (D2a+);
\node[param, above=1mm of D2a] (D2b) {$\text{MLP}_1$};
\coordinate[left=2.5mm of D2b] (D2b_);
\coordinate[right=2.5mm of D2b] (D2b+);
\node[below=-3mm of D2a] (D2c) {$\vdots$};

\node[param, below=4mm of D2a] (D2d) {$\text{MLP}_1$};
\coordinate[left=2.5mm of D2d] (D2d_);
\coordinate[right=2.5mm of D2d] (D2d+);
\node[fixed, right=5mm of D2a] (D3) {\rotatebox{90}{$(n, 1024)$}};
\node[oper, right=2.5mm of D3] (D4) {\rotatebox{90}{bottleneck}}; 
\node[fixed, right=2.5mm of D4] (D5) {\rotatebox{90}{$(1024, )$}};
\coordinate[right=2.5mm of D5] (D6);
\node[param, right=2.5mm of D6] (D7) {\rotatebox{90}{$\text{MLP}_2$}};
\node[fixed, right=2.5mm of D7] (D8) {\rotatebox{90}{$(d,)$}};
\node[oper, right=2.5mm of D8] (D9) {\rotatebox{90}{$\text{LRN}$}};
\node[fixed, right=2.5mm of D9] (D10) {\rotatebox{90}{$\mathbf{f}, (d,)$}};
\coordinate[above=15mm of D6] (E0);
\node[oper, right=5mm of E0] (E1) {\rotatebox{90}{$\lVert \cdot \rVert_{2}$}};
\node[fixed, right=2.5mm of E1] (E2) {\rotatebox{90}{$\rho$}};
\draw[-] (B0) -- (C0);
\draw[-] (C0) -- (C1);
\draw[->] (C1) -- (C2);
\draw[->] (C2) -- (C3);
\draw[-] (C3) -- (C4);
\draw[->] (C4) -- (D0);
\draw[->] (C0) -- (D0);
\draw[->] (D0) -- (D1);
\draw[-] (D1) -- (D2);
\draw[->] (D2) -- (D2a);
\draw[-] (D2) -- (D2b_);
\draw[->] (D2b_) -- (D2b);
\draw[-] (D2) -- (D2d_);
\draw[->] (D2d_) -- (D2d);
\draw[-] (D2a) -- (D2a+);
\draw[-] (D2b) -- (D2b+);
\draw[-] (D2b+) -- (D2a+);
\draw[-] (D2d) -- (D2d+);
\draw[-] (D2d+) -- (D2a+);
\draw[->] (D2a+) -- (D3);
\draw[->] (D3) -- (D4);
\draw[->] (D4) -- (D5);
\draw[-] (D5) -- (D6);
\draw[->] (D6) -- (D7);
\draw[->] (D7) -- (D8);
\draw[->] (D8) -- (D9);
\draw[->] (D9) -- (D10);
\draw[-] (D6) -- (E0);
\draw[->] (E0) -- (E1);
\draw[->] (E1) -- (E2);
\coordinate[right=4mm of D2b] (tmp1);
\coordinate[above=4mm of tmp1] (TR);
\coordinate[left=4mm of D2d] (tmp2);
\coordinate[below=8mm of tmp2] (BL);
\begin{scope}[on background layer]
\node[fill, rectangle, inner sep=0pt, rounded corners, fit={(TR) (BL)}, fill=blue!5] (fit) {};
\coordinate[above=2.5mm of BL] (tmp3);
\node[right=0mm of tmp3, color=blue!40] (text) {\small Shared MLPs};
%
\end{scope} 
\end{tikzpicture}}
  \vspace{-6mm}
  \caption{PointNet-based architecture to encode an input patch $\mathcal{X}$ into a unitary-length $d$-dimensional descriptor $\mathbf{f}$.
  TNet learns the affine transformation $\mathbf{A}$.
  The first Multilayer Perceptron (MLP$_1$) block consists of three shared MLP layers of size (256,512,1024).
  The Bottleneck is a max-pooling layer that produces a 1024-dimension global signature, which is then processed by three MLPs of size (512,256,$d$) (MLP$_2$).
  $\rho$ is the norm of the global signature.
  The Local Response Normalisation (LRN) layer performs a L2 normalisation of the output.
  Except for the last MLP, Batchnorm is used for all layers together with ReLU.
  Dropout is used for the last MLP.
  Colour key: grey = input/output tensors, blue = parametric layer, white = non-parametric layer.}
  \label{fig:net_arch}
\end{figure}
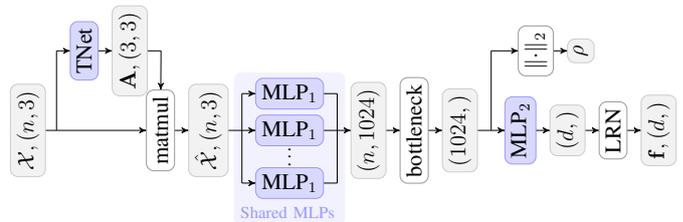

Fig.~\ref{fig:net_arch} shows our PointNet-based architecture \cite{Qi2017a}, where the three main modifications that allow us to produce DIPs are in the Transformation Network, the Bottleneck and the Local Response Normalisation layer.

\vspace{1mm}
\noindent \textbf{Transformation Network} 
The patch $\mathcal{X}$ is firstly passed through the Transformation Network (TNet) that predicts the affine transformation $\mathbf{A} \in \mathbb{R}^{3\times3}$ and that is applied to each $\mathbf{x} \in \mathcal{X}$.
In our experiments we explored the possibility of constraining TNet to be close to an orthogonal matrix, hence a rigid transformation, via the regularisation term $\ell_{\mathrm{reg}} = \lVert \mathbf{I} - \mathbf{A}\mathbf{A}^\top \rVert^2_F$ \cite{Qi2017a,Huynh2009}.
We observed that while $\det(\mathbf{A}) \!\! \rightarrow \!\! 1$, which is a necessary condition for $\mathbf{A}$ to be a rigid transformation, $\mathbf{A} \! \rightarrow \! \mathbf{I}$ as well, thus making the contribution of $\mathbf{A}$ negligible.
We empirically observed performance improvements without constraining TNet to be orthogonal.
There also exists the possibility of using an Iterative Transformer Network, to perform patch canonicalisation iteratively through a series of 3D rigid transformations \cite{Yuan2019}. 
This could be an interesting extension of our architecture, but it is out of the scope of our paper.
TNet has a similar architecture to the rest of the network, except for the last MLP layer that outputs nine values that are opportunely reshaped to form $\mathbf{A}$.
Although TNet is also used in the original version of PointNet \cite{Qi2017a}, we have to train it carefully to achieve the desired behaviour for DIPs.
We explain how we train TNet in Sec.~\ref{sec:loss_functions} and \ref{sec:dataset_training}.
Applying $\mathbf{A}$ to $\mathbf{x}$ results in $\hat{\mathbf{x}} = \mathbf{A} \mathbf{x}$. 
$\hat{\mathcal{X}} = \{ \hat{\mathbf{x}} \}$ is then processed through the MLP layers with shared weights before reaching the bottleneck.

\vspace{.1cm}
\noindent \textbf{Bottleneck} 
The bottleneck is modelled as a symmetric function that produces permutation-invariant outputs \cite{Qi2017a}. 
Although modelling this function with an average pooling operation has shown to be effective for 6DoF registration applications \cite{Wang2019}, we empirically found that max pooling provides superior performance \cite{Qi2017a}. 
Let $m$ be the number of channels in output from the layer before the bottleneck, the max pooling operation is defined as $\max \colon \mathbb{R}^{n \times m} \! \to \! \mathbb{R}^{m}$ such that
\begin{equation}\label{eq:max_pool}
    \boldsymbol{\gamma} = \max_{\mathcal{X}} \Bigl( \Phi_{\boldsymbol{\Theta}_j} (\mathcal{X}) \Bigr),
\end{equation}
where $\boldsymbol{\gamma} = (g_1, g_2, \dots, g_m)$ is a \emph{global signature} of the input patch, $g_i$ is the $i^{\mathrm{th}}$ element of $\boldsymbol{\gamma}$ and $\Phi_{\boldsymbol{\Theta}_j}$ is an intermediate output of the network at the $j^{th}$ layer.
We observed that $\boldsymbol{\gamma}$ can be used to predict how informative DIP is.
Let us define the function that returns the indices of the $\boldsymbol{\gamma}$ values as $\argmax \colon \mathbb{R}^{n \times m} \! \to \! [1,n]^m$ such that 
\begin{equation}\label{eq:argmax_pool}
    \boldsymbol{\alpha} = \argmax_{\mathcal{X}} \Bigl( \Phi_{\boldsymbol{\Theta}_j} (\mathcal{X}) \Bigr).
\end{equation}
Next, let us take two corresponding patches $\mathcal{X}$ and $\mathcal{X}'$ extracted from two overlapping point clouds $\mathcal{P}$ and $\mathcal{P}'$, and then compute $\boldsymbol{\gamma}, \boldsymbol{\alpha}$, and $\boldsymbol{\gamma}', \boldsymbol{\alpha}'$, respectively. 
$\boldsymbol{\alpha}$ ($\boldsymbol{\alpha}'$) will be the same regardless of the permutations of the points in $\mathcal{X}$ ($\mathcal{X}'$).
We observed that the corresponding values of $\boldsymbol{\alpha}$ and $\boldsymbol{\alpha}'$ can be interpreted as the correspondences between points in $\mathcal{X}$ and $\mathcal{X}'$.
Accordingly, their corresponding max values $\boldsymbol{\gamma}$ and $\boldsymbol{\gamma}'$ quantify how reliable these correspondences are.
Then, we found that the norm of $\boldsymbol{\gamma}$ can be effectively used to quantify the reliability of $\boldsymbol{\gamma}$, e.g.~to lower the importance of, or discard, patches extracted from flat surfaces.
It turns out that good descriptors can be selected imposing the condition
\begin{equation}\label{eq:rho}
    \rho = \lVert \boldsymbol{\gamma} \rVert_2 > \tau_\rho,
\end{equation}
where $\tau_\rho$ is a threshold.

\begin{figure}[t]
  \centering
  \includegraphics[width=.8\columnwidth]{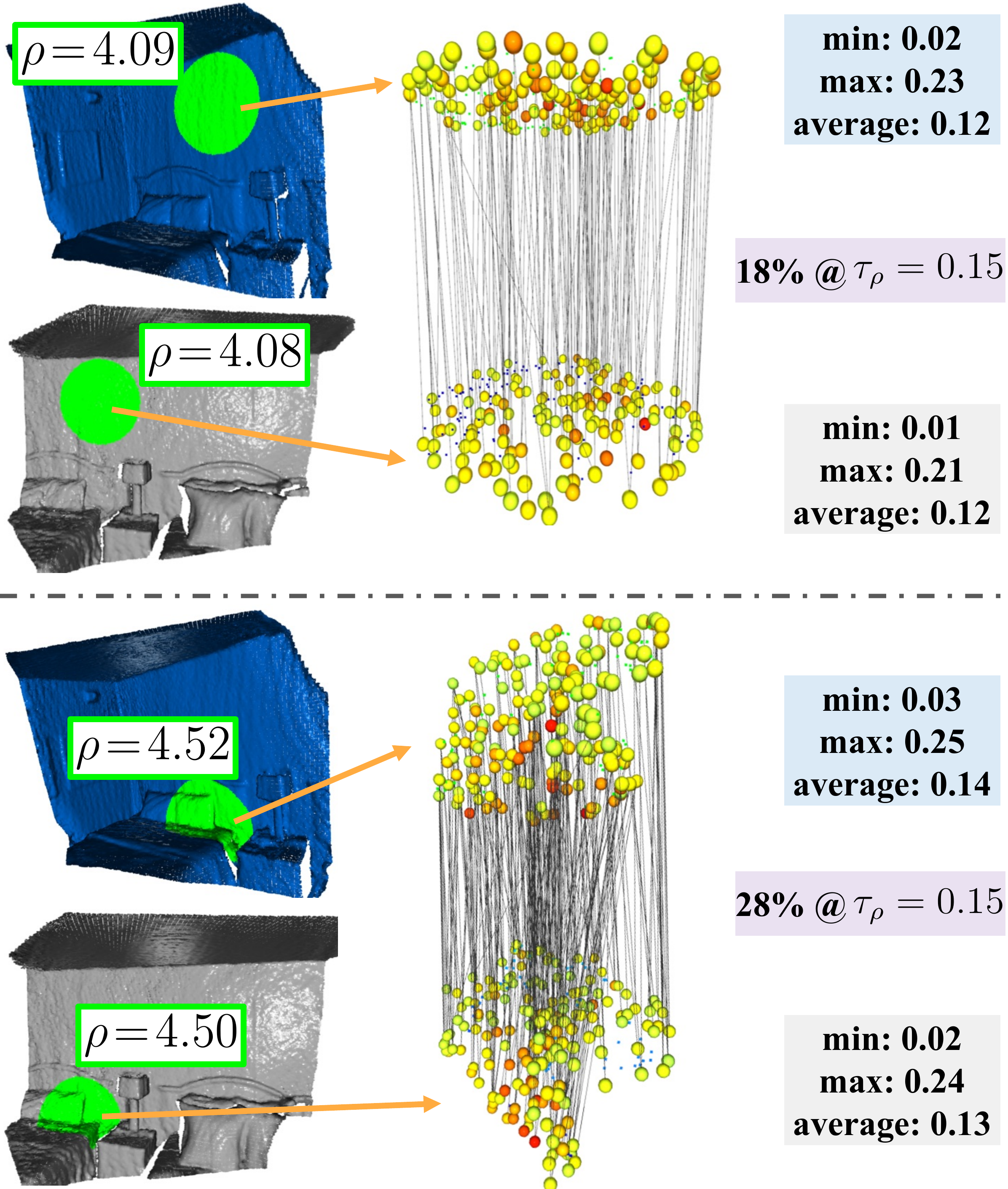}
  \vspace{-3mm}
  \caption{Point correspondences computed from the global signatures of two pairs of corresponding patches (green) that are extracted from two overlapping point clouds (blue and grey) from the 3DMatch dataset \cite{Zeng2017}. 
  (top) Patches extracted from flat surfaces.
  (bottom) Patches extracted from structured surfaces. 
  256 points are randomly sampled from each patch, which are given to our deep neural network as input to produce the global signature.
  Only the correspondences that satisfy the condition $g_i > \tau_\rho = 0.15$ are drawn.
  The percentage of these correspondences is reported.
  Points are colour-coded based on their respective $g_i$ value.
  Minimum, maximum and average $g_i's$ values are reported for each case.}
  \label{fig:glob_signature}
\end{figure}

\begin{figure}[t]
\begin{center}
  \begin{tabular}{@{}c@{}c@{}c}
    \includegraphics[width=.33\columnwidth]{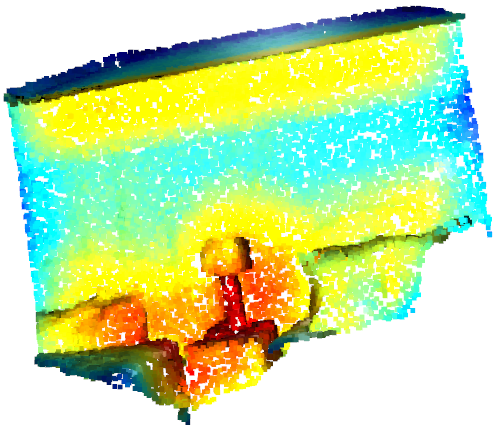}&
    \includegraphics[width=.33\columnwidth]{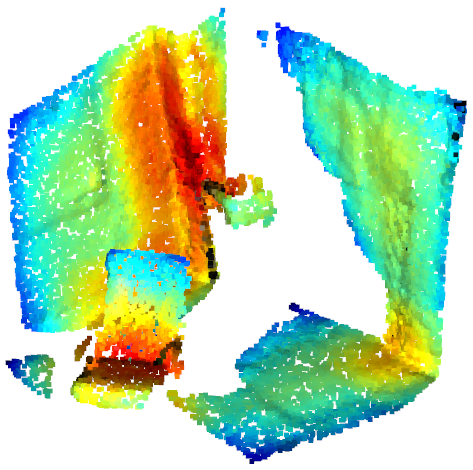}&
    \includegraphics[width=.33\columnwidth]{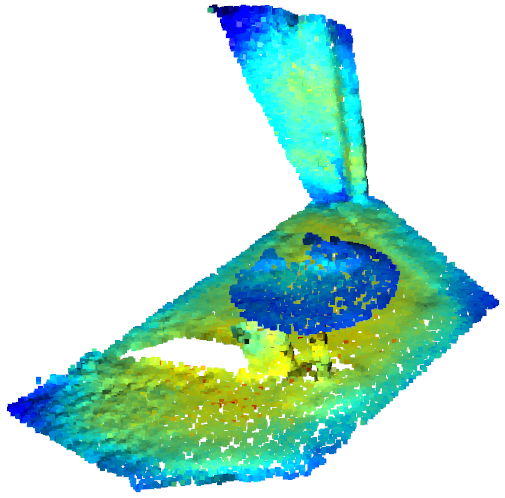}\\
  \end{tabular}
\end{center}
\vspace{-.3cm}
\caption{Heatmaps of the $\rho$ values for 20K patches randomly sampled from three point clouds of the 3DMatch dataset \cite{Zeng2017}. 
Each point is the centre of a patch with a radius of $0.3\sqrt{3}$m \cite{Gojcic2019}.
The more structured the surface enclosed in a patch is, the higher the value of $\rho$ is.}
\label{fig:heatmaps}
\end{figure}

Fig.~\ref{fig:glob_signature} shows an example of global signatures computed from two pairs of corresponding patches (green) that are extracted from two overlapping point clouds (blue and grey) from the 3DMatch dataset \cite{Zeng2017}.
The first case shows two patches extracted from a flat surface (wall), whereas the patches in the second case are extracted from more structured surfaces (bed).
From each patch we randomly sample 256 points and pass them through the network to obtain their respective global signatures (Eq.~\ref{eq:max_pool}).
This figure shows the correspondences between the points of the corresponding patches such that $g_i > \tau_\rho$ $\forall\, i = 1, \dots, m$, where $\tau_\rho = 0.15$.
There are a few things we can observe in this example.
First, values of $\boldsymbol{\gamma}$ are on average higher when the patches are extracted on structured surfaces.
Second, the percentage of correspondences above the threshold is larger when the patches are extracted on structured surfaces ($28\%$ vs. $18\%$).
Lastly, we can see that the patches extracted on flat surfaces have lower $\rho$.
Fig.~\ref{fig:heatmaps} shows the distribution of the $\rho$ values for 20K patches randomly sampled from three point clouds.
We can see that low values of $\rho$ are distributed on flat surfaces (poor information) and along borders (incomplete information).
Differently, $\rho$ has higher value near corners and on objects.

It may be impractical to find a single $\tau_\rho$ that generalises across different input data or different network architecture.
We thereby propose to infer it from the distribution of the $\rho$ values.
Let $\mathcal{R}$ be the set of $\rho$ values computed from the patches extracted from $\mathcal{P}$, and let $f(\rho)$ be the probability density function of the $\rho$ values.
We determine $\tau_\rho$ as the $\mathsf{p}_\rho^{\mathrm{th}}$ percentile through the cumulative density function, such that
\begin{equation}\label{eq:percentile}
    \mathsf{p}_\rho = 100 \cdot \! \int_{-\infty}^{\tau_\rho} f(\rho) d\rho,
\end{equation}
i.e.~the area under the probability density function $f(\rho)$ to the left of $\tau_\rho$ is $\mathsf{p}_\rho / 100$.

\vspace{1mm}
\noindent \textbf{Metric layers and Local Response Normalisation} 
After max pooling, $\boldsymbol{\gamma}$ is processed by a series of MLP layers acting as metric layers to learn distinctive embeddings for our descriptors.
We use a Local Response Normalisation (LRN) layer to produce unitary-length descriptors as we found it works well in practice \cite{Tian2017,Gojcic2019,Choy2019}. 
LRN consists of a L2 normalisation of the last MLP layer's $d$-dimensional output.

\begin{figure*}[t]
  \centering
  \includegraphics[width=1\textwidth]{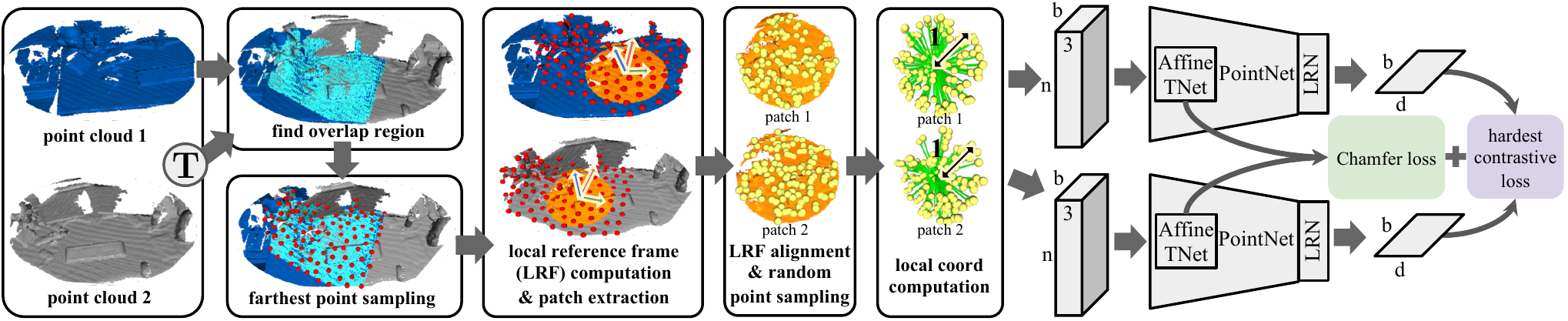}
  \vspace{-6mm}
  \caption{DIP's training pipeline.
  Two overlapping point clouds are aligned using the ground-truth transformation.
  A set of $b$ points (red) belonging to the overlap region (cyan) is sampled using the Farthest Point Sampling method \cite{Qi2017b}.
  We use a Siamese approach to train two deep neural networks with shared parameters concurrently.
  For each branch, we perform the following operations:
  (i) for each point a patch (orange) with radius $\tau_r$ is extracted and the corresponding Local Reference Frame (LRF) \cite{Gojcic2019} is computed using the points of the patch;
  (ii) this patch is rigidly transformed using the LRF and $n$ points are randomly sampled from the patch (yellow points);
  (iii) the coordinates of these $n$ points are expressed relative to the patch centre and normalised in order to have a unitary radius;
  (iv) these $n$ points are given to the deep network as input to learn the descriptor.
  We compute the final loss as the linear combination of the Chamfer loss \cite{Zhao2019} applied to the TNet's output and of the hardest-contrastive loss \cite{Choy2019} applied to the network's output.}
  \label{fig:training_block_diagram}
\end{figure*}

\subsection{Loss functions}\label{sec:loss_functions}

The objective of our training is to produce descriptors whose reciprocal distance in the embedding space is minimised for corresponding patches of different point clouds.
To this end, we train our network following a Siamese approach that processes pairs of corresponding descriptors using two branches with shared weights \cite{Deng2018cvpr,Gojcic2019,Choy2019}.
Each branch independently calculates a descriptor for a given patch.
We learn the parameters of the network by minimising the linear combination of two losses, aiming at two different goals.
The first goal is to geometrically align two patches under the learnt affine transformation.
The second goal is to produce compact and distinctive descriptors via metric learning. 

\vspace{1mm}
\noindent \textbf{Chamfer loss}
Given two patches $\mathcal{X}, \mathcal{X}'$, we want to minimise the distance between each point $\mathbf{x} \in \mathcal{X}$ and its nearest neighbour $\mathbf{x}' \in \mathcal{X}'$. 
Therefore we use the Chamfer loss \cite{Zhao2019,Labrador2020} on the output of TNet as
\begin{align}\label{eq:chamfer_loss}
  \ell_c(\mathcal{X}) = \frac{1}{2n} & \,\,\, \Bigl( \sum_{\mathbf{x} \in \mathcal{X}} \min_{\mathbf{x}' \in \mathcal{X}'} \lVert \mathbf{A}\mathbf{x} - \mathbf{A}'\mathbf{x}' \rVert_2 \Bigr. \\ \nonumber
  & \Bigl. + \sum_{\mathbf{x}' \in \mathcal{X}'} \min_{\mathbf{x} \in \mathcal{X}} \lVert \mathbf{A}\mathbf{x} - \mathbf{A}'\mathbf{x}' \rVert_2 \Bigr).
\end{align}

\vspace{1mm}
\noindent \textbf{Hardest-contrastive loss}
Our metric learning is performed through negative mining using the hardest-contrastive loss \cite{Choy2019}.
Given a pair of anchors $(\mathbf{f}, \mathbf{f}')$, we mine the hardest-negatives $(\mathbf{f}_\text{-}$, $\mathbf{f}_\text{-}')$ and define the loss as 
\begin{align}\label{eq:hardest_contrastive}
    \ell_h = \frac{1}{b} \sum_{(\mathbf{f}, \mathbf{f}') \in \mathcal{C}_\text{+}} & \Bigg( 
    \frac{1}{|\mathcal{C}_\text{+}|} \left[ d(\mathbf{f}, \mathbf{f}') - m_\text{+} \right]_+^2 \Bigr. \\\nonumber
    & \Bigl. + \frac{1}{2 |\mathcal{C}_\text{-}|} [ m_\text{-} - \underbrace{\min_{\tilde{\mathbf{f}} \in \mathcal{C}_\text{-}} d(\mathbf{f}, \tilde{\mathbf{f}})}_{d(\mathbf{f}, \mathbf{f_\text{-}})} ]_+^2 \Bigr. \\\nonumber
    & \Bigl. + \frac{1}{2 |\mathcal{C}_\text{-}|} [ m_\text{-} - \underbrace{\min_{\tilde{\mathbf{f}} \in \mathcal{C}_\text{-}} d(\mathbf{f}', \tilde{\mathbf{f}})}_{d(\mathbf{f'}, \mathbf{f'_\text{-}})} ]_+^2 \Bigg), \\\nonumber
\end{align}
where $\mathcal{C}_\text{+}$ is the set of the anchor pairs and $\mathcal{C}_\text{-}$ is the set of descriptors (opportunely sampled) used for the hardest-negative mining extracted from a minibatch.
$m_\text{+}$ and $m_\text{-}$ are the margins for positive and negative pairs, respectively. $[\cdot]_+$ takes the positive part of its argument.

\section{Experimental validation}

We evaluate the distinctiveness of DIPs using the indoor 3DMatch dataset \cite{Zeng2017}, and assess DIP generalisation on the outdoor ETH dataset \cite{Pomerleau2012} and on a new indoor dataset we collected with a smartphone.
We explain how patches are extracted and given as input to our deep network.
The training pipeline is shown in Fig.~\ref{fig:training_block_diagram}.
Our method is developed in Pytorch 1.3.1 \cite{Pytorch}.
We compare our method with 14 state-of-the-art methods and carry out a thorough ablation study.

\subsection{Patch extraction}\label{sec:patch_extraction}

DIPs are learnt from patches that are extracted from point cloud pairs $(\mathcal{P},\mathcal{P}')$ whose overlap region is greater than a threshold $\tau_o$.
Let $\mathcal{O} \! \subset \! \mathcal{P}$ and $\mathcal{O}' \! \subset \! \mathcal{P}'$ be the overlap regions.
During training we know the ground-truth transformation $\mathbf{T} \! \in \! SE(3)$ that register $\mathcal{P}'$ to $\mathcal{P}$.
Point correspondences between $\mathcal{O}$ and $\mathcal{O}'$ can be determined either by using the 3DMatch toolbox \cite{Zeng2017}, or by using a nearest neighbourhood search after applying $\mathbf{T}$ to $\mathcal{P}'$ \cite{Gojcic2019}.
We use the latter approach by seeking nearest points from $\mathcal{O}$ to $\mathcal{O}'$ within a radius of $10cm$.
Corresponding points in $\mathcal{O}$ and $\mathcal{O}'$ are the candidate anchors used by the hardest-contrastive loss (Eq.~\ref{eq:hardest_contrastive}).

\vspace{.1cm}
\noindent \textbf{Farthest Point Sampling} 
Anchor sampling is key to allow for an effective minimisation of Eq.~\ref{eq:hardest_contrastive}, and typically this is carried out with random sampling \cite{Gojcic2019,Choy2019}.
Such random sampling may lead to cases where anchors and negatives are sampled spatially close to each other.
A solution can be disregarding negatives within a certain radius from an anchor by computing the Euclidean distances amongst all the anchors within a minibatch in order to determine whether to penalise for the distance between descriptors in the embedding space (Eq.~5 in \cite{Choy2019}).
Including points within the radius would force the network to learn distinctive descriptors of region with similar geometric structures, thus making training unstable.
Therefore to avoid computing the Euclidean distances amongst all the anchors within the minibatch \cite{Choy2019,Gojcic2019}, we efficiently sample anchors having the largest distance amongst themselves using Farthest Point Sampling (FPS) \cite{Qi2017b}.
Specifically, we sample $b$ points within $\mathcal{O}$ using FPS and then search for the nearest neighbour counterparts in $\mathcal{O}'$.
These points are the anchors that construct the minibatch on which the hardest-negative mining is performed.
In our experiments we use $b=256$.

\vspace{1mm}
\noindent \textbf{Local patch}
Given a point $\mathbf{c} \in \mathcal{O}$ sampled with FPS, and its nearest neighbour $\mathbf{c}' \in \mathcal{O}'$, we build the set $\mathcal{Y} = \{ \mathbf{y} \in \mathcal{X} : \lVert \mathbf{y} - \mathbf{c} \rVert_2 \le \tau_r \}$, and similarly the set $\mathcal{Y}'$ for $\mathbf{c}'$.
$\tau_r$ is the radius of our patch.
We use the points in $\mathcal{Y}$ and $\mathcal{Y}'$ to compute their own Local Reference Frame (LRF) \cite{Yang2016,Gojcic2019}.
Each LRF is constructed independently by computing the three orthogonal axes: 
the z-axis is computed as the normal of the local surface defined by the points of $\mathcal{Y}$ ($\mathcal{Y}'$); 
the x-axis is computed as a weighted sum of the vectors constructed as the projection of vectors between $\mathbf{c}$ and the points in $\mathcal{Y} \! \setminus \! \mathbf{c}$ on the plane orthogonal to the z-axis;
the y-axis is computed as the cross-product between the z-axis and the x-axis.
We implement the LRF following \cite{Gojcic2019}.
Let $\mathbf{L}, \mathbf{L}' \in \mathbb{R}^{3 \times 3}$ be the LRFs of $\mathcal{Y}$ and $\mathcal{Y}'$, respectively.
$\mathcal{Y}$ and $\mathcal{Y}'$ may contain a large number of points, typically a few thousands, thus it is impractical to process them all with a deep network.
Similarly to \cite{Qi2017a}, we randomly sample $n\!=\!256$ points.
This also helps regularisation during training and generalisation.
Next, we recalculate the coordinate of each of the $n$ points relative to their patch centre and normalise the radius of the sphere that contains them.
Formally, let $Q(\mathcal{Y}) = \{ \hat{\mathbf{y}} \colon \hat{\mathbf{y}} = (\mathbf{y} - \mathbf{c}) / \tau_r, \mathbf{y} \in \mathcal{Y} \}$ be the set of randomly-sampled and normalised points from $\mathcal{Y}$ where $\lvert Q(\mathcal{Y}) \rvert = n$. 
Lastly, we apply $\mathbf{L}$ to $Q(\mathcal{Y})$ to rotate the points with respect to their LRFs, such that
\begin{equation}
    \mathcal{X} = \mathbf{L} \otimes Q(\mathcal{Y}),
\end{equation}
where the operation $\otimes$ defines the application of $\mathbf{L}$ to each element of $Q(\mathcal{Y})$ such that $\mathbf{x} = \mathbf{L} \hat{\mathbf{y}}$.
Analogously, the same operations are performed for $Q(\mathcal{Y}')$.

\subsection{Datasets, training and testing setup}\label{sec:dataset_training}

\noindent \textbf{3DMatch dataset} 
We learn DIPs from the point clouds of the 3DMatch dataset \cite{Zeng2017}.
3DMatch is composed of 62 real-world indoor scenes collected from Analysis-by-Synthesis \cite{Valentin2016}, 7-Scenes \cite{Shotton2013}, SUN3D \cite{Xiao2013}, RGB-D Scenes v.2 \cite{Kim2014}, and Halber and Funkhouser \cite{Halber2016}.
The official split consists of 54 scenes for training and 8 for testing.
Each scene is split into partially overlapping and registered point cloud pairs.
As in \cite{Gojcic2019}, we train our deep network with the pairs whose overlap is more than $\tau_o \! = \! 30\%$.
The $b=256$ points are sampled from each of these overlap regions and we centre the patches on these points to construct each minibatch.
As in \cite{Choy2019}, we set $m_\text{+} = 0.1$ and $m_\text{-} = 1.4$ (Eq.~\ref{eq:hardest_contrastive}).
Each epoch consists of 16602 iterations.
Each iteration is for a point cloud pair.
We train for 40 epochs.
We use Dropout with probability $0.3$ at the last MLP layer.
We subsample point clouds using a voxel size of $0.01$m.
As \cite{Gojcic2019}, we set $\tau_r = 0.3\sqrt{3}$m.
Our training aims to minimise the linear combination of $\ell_h$ (Eq.~\ref{eq:hardest_contrastive}) and $\ell_c$ (Eq.~\ref{eq:chamfer_loss}) as
\begin{equation}
    \ell = \ell_h + \frac{1}{b} \sum_{\mathcal{X} \in \mathcal{P}} \ell_c(\mathcal{X}).
\end{equation}

We use Stochastic Gradient Descent with an initial learning rate of $10^{-3}$ that decreases by a factor $0.1$ every 15 epochs.
The eight test scenes consists of 1117 point cloud pairs.
As in \cite{Gojcic2019,Choy2019,Bai2020}, testing is performed by randomly sampling 5K points from each point cloud.
To evaluate DIP's rotation invariance ability, we follow the evaluation of \cite{Gojcic2019} and create an augmented version of 3DMatch, namely 3DMatchRotated: each point cloud is rotated by an angle sampled uniformly between $[0,2\pi]$ around all the three axes independently.
Unless otherwise stated we use $\mathsf{p}_\rho = 5$.

\vspace{1mm}
\noindent \textbf{ETH dataset} 
We use the ETH dataset to assess the ability of DIPs to generalise across sensor modalities (RGB-D $\!\rightarrow\!$ laser scanner) and on different scenes (indoor $\!\rightarrow\!$ outdoor) \cite{Pomerleau2012}.
To this end we use the same model trained on the 3DMatch dataset (no fine tuning).
The ETH dataset consists of four outdoor scenes, namely \emph{Gazebo-Summer}, \emph{Gazebo-Winter}, \emph{Wood-Summer} and \emph{Wood-Autumn}, containing partially overlapping, sparse and dense vegetation point clouds.
Differently from the 3DMatch dataset we subsample point clouds using a voxel size of $0.06$m.
We set the patch kernel size $\tau_r = 0.6\sqrt{3}$m.
For a fair comparison, the evaluation procedure follows verbatim \cite{Gojcic2019}, i.e.~random sampling of 5K points.

\vspace{1mm}
\noindent \textbf{VigoHome dataset} 
To evaluate DIPs on another sensor modality (RGB-D $\!\rightarrow\!$ smartphone RGB), we have created a new dataset, namely VigoHome, by reconstructing the inside of a house using a Visual-SLAM smartphone App we developed with ARCore (Android) \cite{arcore}.
We captured three zones, namely \emph{livingroom-downstairs} (94K points), \emph{bedroom-upstairs} (43K points), and \emph{bathroom-upstairs} (83K points).
The stairs between the three zones is the overlapping region of the point clouds.
We calculated their transformations to a common reference frame and determined the point correspondences to evaluate the registration:
two points of a point cloud pair are corresponding if they are nearest neighbours within a 0.1m-radius.
We subsample point clouds using voxels of 0.01m and set $\tau_r=0.6\sqrt{3}$m.

\subsection{Comparison and ablation study setup}

We compare DIPs against 14 alternative descriptors: Spin~\cite{Johnson1999}, SHOT~\cite{Salti2014}, FPFH~\cite{Rusu2009}, USC~\cite{Tombari2010}, CGF~\cite{Khoury2017}, 3DMatch~\cite{Zeng2017}, Folding~\cite{Yang2017}, PPFNet~\cite{Deng2018cvpr}, PPF-FoldNet~\cite{Deng2018eccv}, DirectReg~\cite{Deng2019}, CapsuleNet~\cite{Zhao2019}, PerfectMatch~\cite{Gojcic2019}, FCGF~\cite{Choy2019}, and D3Feat~\cite{Bai2020}.
In our ablation study we train the model for five epochs on a subset of 3DMatch's scenes, i.e.~\emph{Chess} and \emph{Fire}, and test on \emph{Home2} and \emph{Hotel3}.
We chose these test scenes because we found them to be sufficiently challenging.

\subsection{Evaluation metrics}

\noindent \textbf{Feature-matching recall}
We use the feature-matching recall (FMR) to quantify the descriptor quality \cite{Deng2018cvpr}.
FMR does not require RANSAC \cite{Fischler1981} as it directly averages the number of correctly matched point clouds across datasets. 
Only recall is measured, as the precision can be improved by pruning correspondences \cite{Choi2015,Deng2018cvpr}.
FMR is defined as
\begin{equation}
    \Xi = \! \frac{1}{|\mathcal{F}|} \! \sum_{s=1}^{|\mathcal{F}|} \mathds{1} \Bigl( \underbrace{\Bigl[ \frac{1}{|\Omega_s|} \!\! \sum_{(\mathbf{x},\mathbf{x}') \in \Omega_s} \!\!\!\!\!\! \mathds{1} (\lVert \mathbf{x} - \mathbf{T}_s \mathbf{x}' \rVert_2 \! < \! \tau_1) \Bigr]}_{\xi_{\Omega_s}} \! > \! \tau_2 \Bigr),
\end{equation}
where $|\mathcal{F}|$ is the number of matching point cloud pairs having $\tau_o \ge 30\%$ (overlap between each other).
$(\mathbf{x}, \mathbf{x}')$ is a pair of corresponding points found in the descriptor (embedding) space via a mutual nearest-neighbour search \cite{Gojcic2019}.
$\Omega_s$ is the set that contains all the found pairs $(\mathbf{x}, \mathbf{x}')$ in the overlap regions $\mathcal{O} \! \subset \! \mathcal{P}$ and $\mathcal{O}' \! \subset \! \mathcal{P}'$, respectively.
$\mathbf{T}_s$ is the ground-truth transformation alignment between $\mathcal{P}'$ and $\mathcal{P}$.
$\mathds{1}(\cdot)$ is the indicator function.
$\tau_1 = 10$cm and $\tau_2 = 0.05$ are set based on the theoretical analysis that RANSAC will find at least three corresponding points that can provide the correct $\mathbf{T}_s$ with probability 99.9\% using no more than $\approx55K$ iterations \cite{Deng2018cvpr,Gojcic2019}.
In addition to $\Xi$, we also report mean ($\mu_\xi$) and standard deviation ($\sigma_\xi$) of $\xi_{\Omega_s}$ before applying $\tau_2$.

\vspace{1mm}
\noindent \textbf{Registration recall}
We measure the registration recall for the transformation estimated with RANSAC \cite{Choy2019}.
The registration recall quantifies the miss-rate by measuring the distance between corresponding points for each point cloud pair using the estimated transformation based on ground-truth point correspondence information.
The registration recall is defined as
\begin{equation}
    \Upsilon = \frac{1}{|\mathcal{F}|} \sum_{s=1}^{|\mathcal{F}|} \mathds{1} \Bigg( \sqrt{\frac{1}{|\Omega_s|} \sum_{(\mathbf{x}, \mathbf{x}') \in \Omega_s} \!\!\!\! \lVert \mathbf{x} - \tilde{\mathbf{T}} \mathbf{x}' \rVert_2^2 }  \! < \! 0.2\mathrm{m} \Bigg),
\end{equation}
where $\tilde{\mathbf{T}}$ is the estimated transformation. 
Only point clouds pairs with at least 30\% overlap are evaluated.
0.2m is set as in \cite{Zeng2017,Choy2019,Bai2020}.
We configure RANSAC based on the FMR formulation and use its Open3D implementation \cite{Zhou2018}.

\subsection{Quantitative analysis and comparison}

\begin{figure}[t]
\begin{center}
  \begin{tabular}{@{}c@{}c}
      \includegraphics[width=.5\columnwidth]{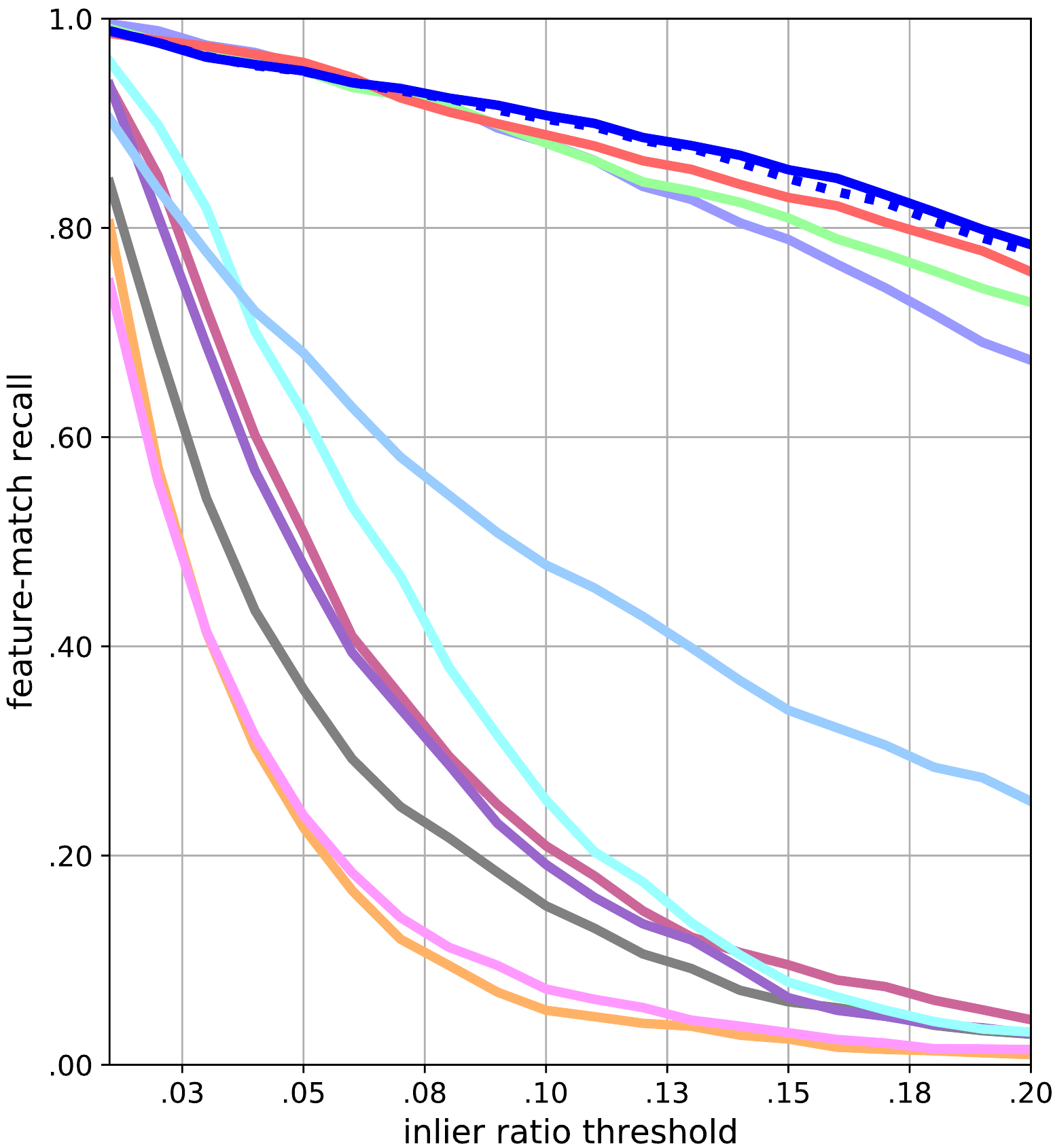}&
      \includegraphics[width=.5\columnwidth]{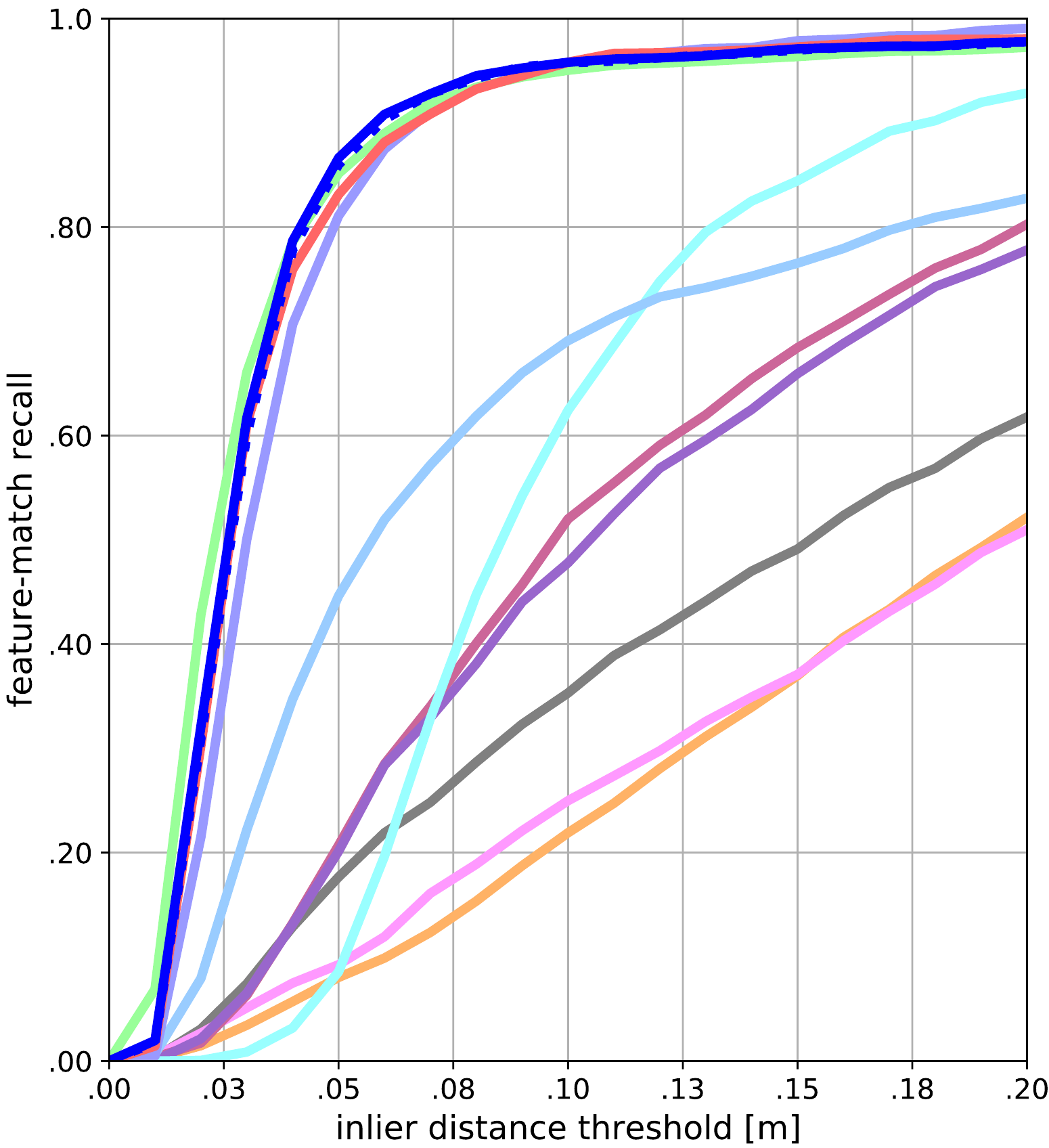}
  \end{tabular}
  \begin{tabular}{@{}c}
    \includegraphics[width=1\columnwidth]{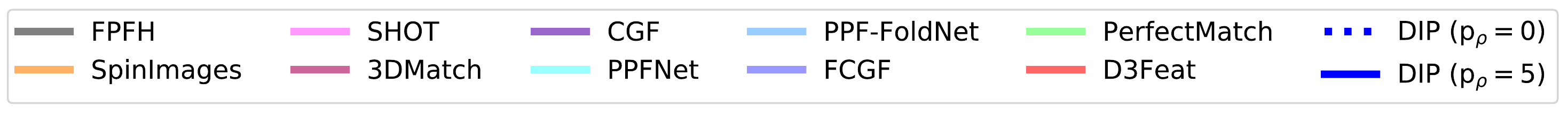}
  \end{tabular}
\end{center}
\vspace{-.3cm}
\caption{Feature-matching recall as a function of (left) $\tau_2$ and (right) $\tau_1$.}
\label{fig:inlier_graphs}
\end{figure}

Tab.~\ref{tab:fmr_3dmatch} reports DIP results in comparison with alternative descriptors on 3DMatch and 3DMatchRotated datasets.
Results show that DIPs achieve state-of-the art results and that are rotation invariant as FMR is almost the same for both datasets.
When FMR is measured at different values of $\tau_2$, DIPs are more distinctive than the alternatives (Fig.~\ref{fig:inlier_graphs}).
Interestingly, DIPs largely outperform PPFNet descriptors that are also computed with a PointNet-based backbone.
We believe that this occurs on the one hand as a result of DIP's LRF canonicalisation, in fact PPFNet's FMR drops in 3DMatchRotated as no canonicalisation is performed, and on the other hand because DIPs encode only the local geometric information, which makes them more generic and distinctive across different scenes as opposed to PPFNet descriptors that instead encode contextual information too.
Amongst the eight tested scenes, the worst performing one is \emph{Lab} that will analyse in detail later.
Our Python implementation takes $4.87$ms to process a DIP using an i7-8700 CPU at 3.20GHz with a NVIDIA GTX 1070 Ti GPU and 16GB RAM. 
93\% of the execution time is for the LRF estimation.
Once a patch is canonicalised the deep network processes the descriptor in $0.37$ms.
Note that the LRF estimation can be parallelised and implemented in C++ to reduce the execution time.

\begin{table}[t]
    \caption{Feature-matching recall on the 3DMatch dataset \cite{Zeng2017}.}
    \vspace{-.2cm}
    \label{tab:fmr_3dmatch}
    \resizebox{\columnwidth}{!}{%
    \begin{tabular}{lcccccc}
        \toprule
        \multirow{2}{*}{Method} & \multicolumn{2}{c}{3DMatch} & \multicolumn{2}{c}{3DMatchRotated} & Feat. & Time \\
        & $\Xi$  & std  & $\Xi$  & std  & dim. & [ms] \\
        \toprule
        Spin~\cite{Johnson1999}             & .227 & .114 & .227 & .121 & 153  & .133   \\
        SHOT~\cite{Salti2014}               & .238 & .109 & .234 & .095 & 352  & .279   \\
        FPFH~\cite{Rusu2009}                & .359 & .134 & .364 & .136 & 33   & .032   \\
        USC~\cite{Tombari2010}              & .400 & .125 & -    & -    & 1980 & 3.712  \\
        CGF~\cite{Khoury2017}               & .582 & .142 & .585 & .140 & 32   & 1.463  \\
        3DMatch~\cite{Zeng2017}             & .596 & .088 & .011 & .012 & 512  & 3.210  \\
        Folding~\cite{Yang2017}             & .613 & .087 & .023 & .010 & 512  & .352   \\
        PPFNet~\cite{Deng2018cvpr}          & .623 & .108 & .003 & .005 & 64   & 2.257  \\
        PPF-FoldNet~\cite{Deng2018eccv}     & .718 & .105 & .731 & .104 & 512  & .794   \\
        DirectReg~\cite{Deng2019}           & .746 & .094 & -    & -    & 512  & .794   \\
        CapsuleNet~\cite{Zhao2019}          & .807 & .062 & .807 & .062 & 512  & 1.208  \\
        PerfectMatch~\cite{Gojcic2019}      & .947 & .027 & .949 & .024 & 32   & 5.515  \\
        FCGF~\cite{Choy2019}                & .952 & .029 & .953 & .033 & 32   & .009   \\
        D3Feat~\cite{Bai2020}               & \textbf{.958} & .029 & \textbf{.955} & .035 & 32 & - \\
        \midrule
        DIP                                 & .948 & .046 & .946 & .046 & 32 & 4.870   \\
        \bottomrule
    \end{tabular}%
    }
\end{table}

\begin{figure}[t]
  \centering
  \includegraphics[width=1\columnwidth]{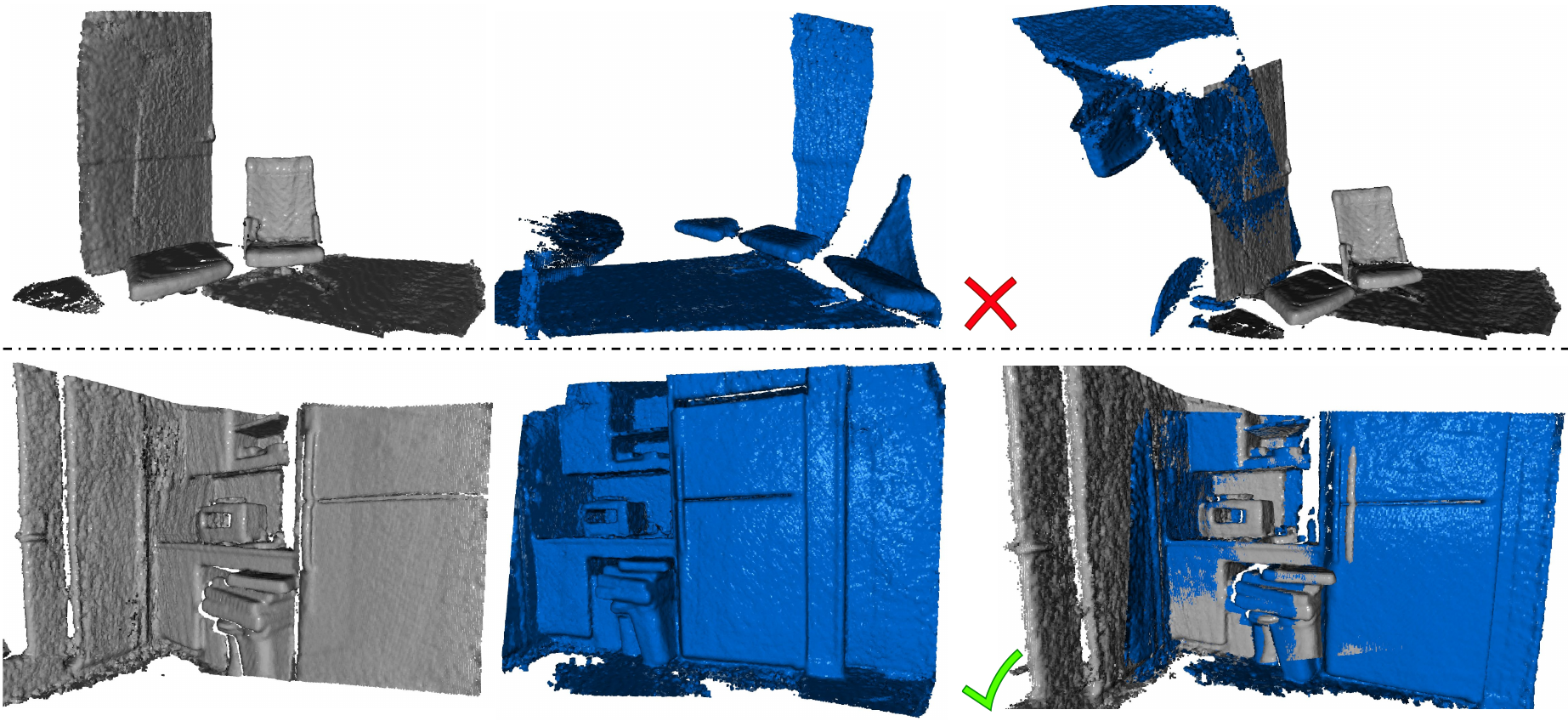}
  \caption{Estimated rigid transformations of two point cloud pairs from the 3DMatch dataset \cite{Zeng2017}. 
  (top) Incorrectly estimated transformation due to the lack of structured surfaces. 
  (bottom) Correctly estimated transformation thanks to the structured elements given by the kitchen appliances.}
  \label{fig:3dmatch_qualitative}
\end{figure}

Tab.~\ref{tab:registration_recall} reports the registration recall results, where the descriptor distinctiveness we have observed in Fig.~\ref{fig:inlier_graphs} is reflected on the estimated transformations.
On average, DIPs outperform all the other descriptors.
We can see that the \emph{Lab} scene mentioned before is the worst performing one.
This occurs because \emph{Lab} contains several point clouds of partially reconstructed objects and flat surfaces.
Two examples are shown in Fig.~\ref{fig:3dmatch_qualitative}.
The first one is a failed registration due to the lack of informative geometries in the scene.
The second one is a successful registration, where the kitchen appliances produced more distinctive descriptors than the first case.

\begin{table}[t]
    \tabcolsep 1pt
    \caption{Registration recall on the 3DMatch dataset \cite{Zeng2017}.}
    \vspace{-.2cm}
    \label{tab:registration_recall}
    \resizebox{1\columnwidth}{!}{%
    \begin{tabular}{lccccccccc}
        \toprule
        Method & Kitchen & Home1 & Home2 & Hotel1 & Hotel2 & Hotel3 & Study & Lab &  Average \\
        \toprule
        FPFH \cite{Rusu2009}        & .36 & .56 & .43 & .29 & .36 & .61 & .31 & .31 & .40 \\
        USC \cite{Tombari2010}      & .52 & .35 & .47 & .53 & .20 & .38 & .46 & .49 & .43 \\
        CGF \cite{Khoury2017}       & .72 & .69 & .46 & .55 & .49 & .65 & .48 & .42 & .56 \\
        3DMatch \cite{Zeng2017}     & .85 & .78 & .61 & .79 & .59 & .58 & .63 & .51 & .67 \\
        PPFNet \cite{Deng2018cvpr}  & .90 & .58 & .57 & .75 & .68 & .88 & .68 & .62 & .71 \\
        FCGF \cite{Choy2019}        & .93 & .91 & .71 & .91 & .87 & .69 & .75 & \textbf{.80} & .82 \\
        \midrule
        DIP                         & \textbf{.98} & \textbf{.94} & \textbf{.85} & \textbf{.98} & \textbf{.92} & \textbf{.89} & \textbf{.80} & .75 & \textbf{.89} \\
        \bottomrule
    \end{tabular}
    }
\end{table}

We further evaluate the registration recall and assess DIP robustness following the comparative ablation study proposed in \cite{Bai2020}, where the registration recall is measured as a function of a decreasing number of sampled points used by RANSAC to estimate the transformation.
Tab.~\ref{tab:comparative_abl} shows that DIPs on average have a superior robustness than the alternatives.

\begin{table}[t]
    \centering
    \caption{Ablation study using the registration recall as a function of the number of sampled points on the 3DMatch dataset \cite{Zeng2017}.}
    \vspace{-.2cm}
    \label{tab:comparative_abl}
    \resizebox{1\columnwidth}{!}{%
    \begin{tabular}{lcccccc}
        \toprule
        \multirow{2}{*}{Method}         &  \multicolumn{5}{c}{\# sampled points} & \multirow{2}{*}{Average}\\
                                        & 5000 & 2500 & 1000 & 500 & 250 & \\
        \toprule
        PerfectMatch~\cite{Gojcic2019}  & .803 & .775 & .734 & .648 & .509 & .694 \\
        FCGF~\cite{Choy2019}            & .873 & .858 & .858 & .810 & .730 & .826 \\
        D3Feat~\cite{Bai2020}           & .822 & .844 & .849 & .825 & \textbf{.793} & .827 \\
        \midrule
        DIP                             & \textbf{.889} & \textbf{.890} & \textbf{.878} & \textbf{.866} & .774 & \textbf{.859} \\
        \bottomrule
    \end{tabular}
    }
\end{table}

\subsection{Ablation study}\label{sec:comparative_abl_study}

Tab.~\ref{tab:abl_study} reports the results of our ablation study on the implementation choices.
Here we can see how the three modules, i.e.~TNet, LRF and LRN, affect FMR.
TNet learns to compensate for incorrectly estimated LRFs.
But we can see that without LRF, TNet cannot learn the complete transformation to canonicalise the patches (FMR drops on 3DMatchRotated).
LRF is key to make DIPs rotation invariant.
Following \cite{Gojcic2019} and \cite{Choy2019}, we can notice how learning unitary-length descriptors improve FMR.
Lastly, as expected, the more the capacity to encode the information in descriptors of larger dimension, the better the performance.
However, we used 32-dimensional descriptors throughout all the experiments in order to compare results with existing descriptors.

\begin{table}[t]
    \tabcolsep 5pt
    \caption{Ablation study on DIP's implementation choices.}
    \vspace{-.2cm}
    \label{tab:abl_study}
    \resizebox{1\columnwidth}{!}{%
    \begin{tabular}{cccccccccc}
        \toprule
        \multirow{2}{*}{$d$} & \multirow{2}{*}{TNet} & \multirow{2}{*}{LRF} & \multirow{2}{*}{LRN} & \multicolumn{3}{c}{3DMatch} & \multicolumn{3}{c}{3DMatchRotated} \\
                & & & & $\Xi$ & $\mu_\xi$ & $\sigma_\xi$ & $\Xi$ & $\mu_\xi$ & $\sigma_\xi$ \\
        \toprule
        32 & & \cmark & \cmark & .886 & .272 & .207 & .877 & .271 & .207 \\
        32 &\cmark & \cmark & & .831 & .215 & .175 & .834 & .214 & .174 \\
        32 &\cmark & & \cmark & .824 & .237 & .195 & .215 & .039 & .054 \\
        \midrule
        16 &\cmark & \cmark & \cmark & .903 & .264 & .193 & .906 & .264 & .193 \\
        32 &\cmark & \cmark & \cmark & .919 & .318 & .218 & .926 & .317 & .217 \\
        64 &\cmark & \cmark & \cmark & .916 & .327 & .221 & .928 & .325 & .220 \\
        128 &\cmark & \cmark & \cmark & .933 & .335 & .222 & .924 & .334 & .222 \\
        \bottomrule
    \end{tabular}
    }
\end{table}

\subsection{Generalisation ability: comparison and analysis}

\begin{figure}[t]
  \centering
  \includegraphics[width=1\columnwidth]{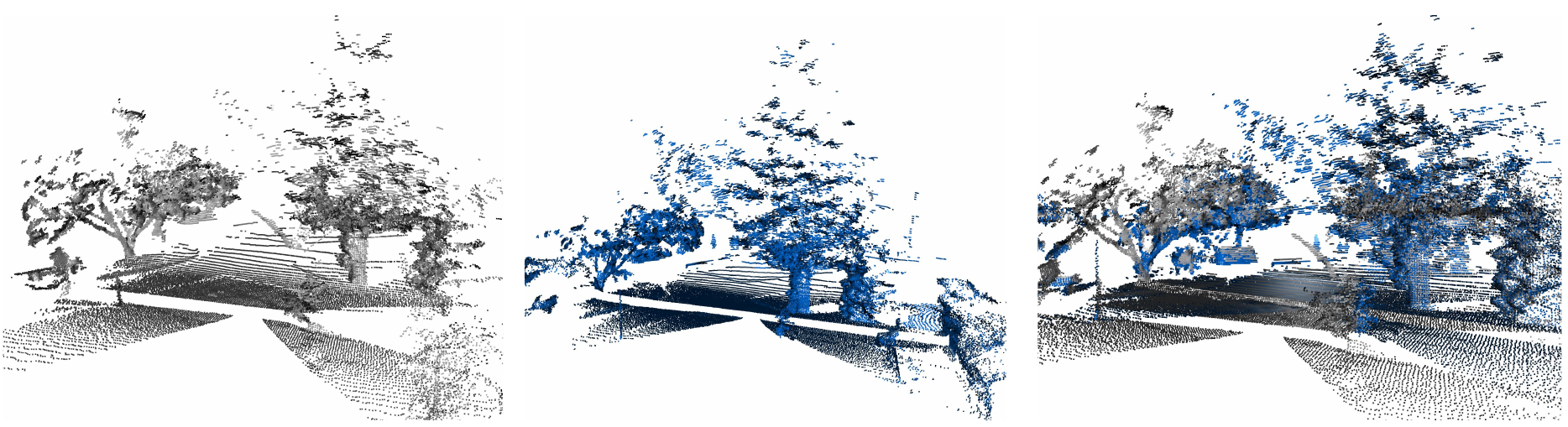}
  \vspace{-6mm}
  \caption{Estimated rigid transformation of a point cloud pair from the ETH dataset \cite{Pomerleau2012}.
  This example shows that DIPs learnt on the 3DMatch dataset (RGB-D reconstructions) can be successfully used to register point clouds reconstructed with a laser scanner in outdoor scenes.}
  \label{fig:eth_qualitative}
\end{figure}

Tab.~\ref{tab:eth_dataset} reports the results obtained on the ETH dataset \cite{Pomerleau2012}.
We can see that DIPs on average largely outperform the alternative descriptors.
Second to DIPs are PerfectMatch's descriptors \cite{Gojcic2019} that, as DIPs, use LRF canonicalisation.
However, differently from DIPs, PerfectMatch's descriptors are learnt from hand-crafted representations, namely \emph{voxelised smoothed density value} \cite{Gojcic2019}.
We argue that letting the network learn the encoding from the points directly (end-to-end), leads to a much greater robustness and generalisation ability.
We can also observe that the application of $\mathsf{p}_\rho$ improves the performance.
Fig.~\ref{fig:eth_qualitative} shows an example of result from Gazebo-Summer.
Although the sensor modality and the structure of the environment is very different from that of 3DMatch, DIPs maintain their distinctiveness and can be successfully used to register two point clouds reconstructed with a laser scanner.

\begin{table}[t]
    \tabcolsep 5pt
    \centering
    \caption{Feature-matching recall on the ETH dataset \cite{Pomerleau2012}.}
    \vspace{-.2cm}
    \label{tab:eth_dataset}
    \resizebox{1\columnwidth}{!}{%
    \begin{tabular}{lccccc}
        \toprule
        \multirow{2}{*}{Method} & \multicolumn{2}{c}{Gazebo} & \multicolumn{2}{c}{Wood} & \multirow{2}{*}{Average} \\ 
        & Summer & Winter & Autumn & Summer & \\
        \toprule
        FPFH~\cite{Rusu2009}            & .386 & .142 & .148 & .208 & .221 \\ 
        SHOT~\cite{Salti2014}           & .739 & .457 & .609 & .640 & .611 \\
        3DMatch~\cite{Zeng2017}         & .228 & .083 & .139 & .224 & .169 \\
        CGF~\cite{Khoury2017}           & .375 & .138 & .104 & .192 & .202 \\
        PerfectMatch~\cite{Gojcic2019}  & \textbf{.913} & .841 & .678 & .728 & .790 \\
        FCGF~\cite{Choy2019}            & .228 & .100 & .148 & .168 & .161 \\
        D3Feat~\cite{Bai2020}           & .859 & .630 & .496 & .480 & .563 \\
        \midrule
        DIP ($\mathsf{p}_\rho=0$)       & .897 & .869 & .957 & .944 & .916 \\
        DIP ($\mathsf{p}_\rho=5$)       & .908 & \textbf{.886} & \textbf{.965} & \textbf{.952} & \textbf{.928} \\
        \bottomrule
    \end{tabular}
    }
\end{table}

\begin{figure}[t]
  \centering
  \includegraphics[width=1\columnwidth]{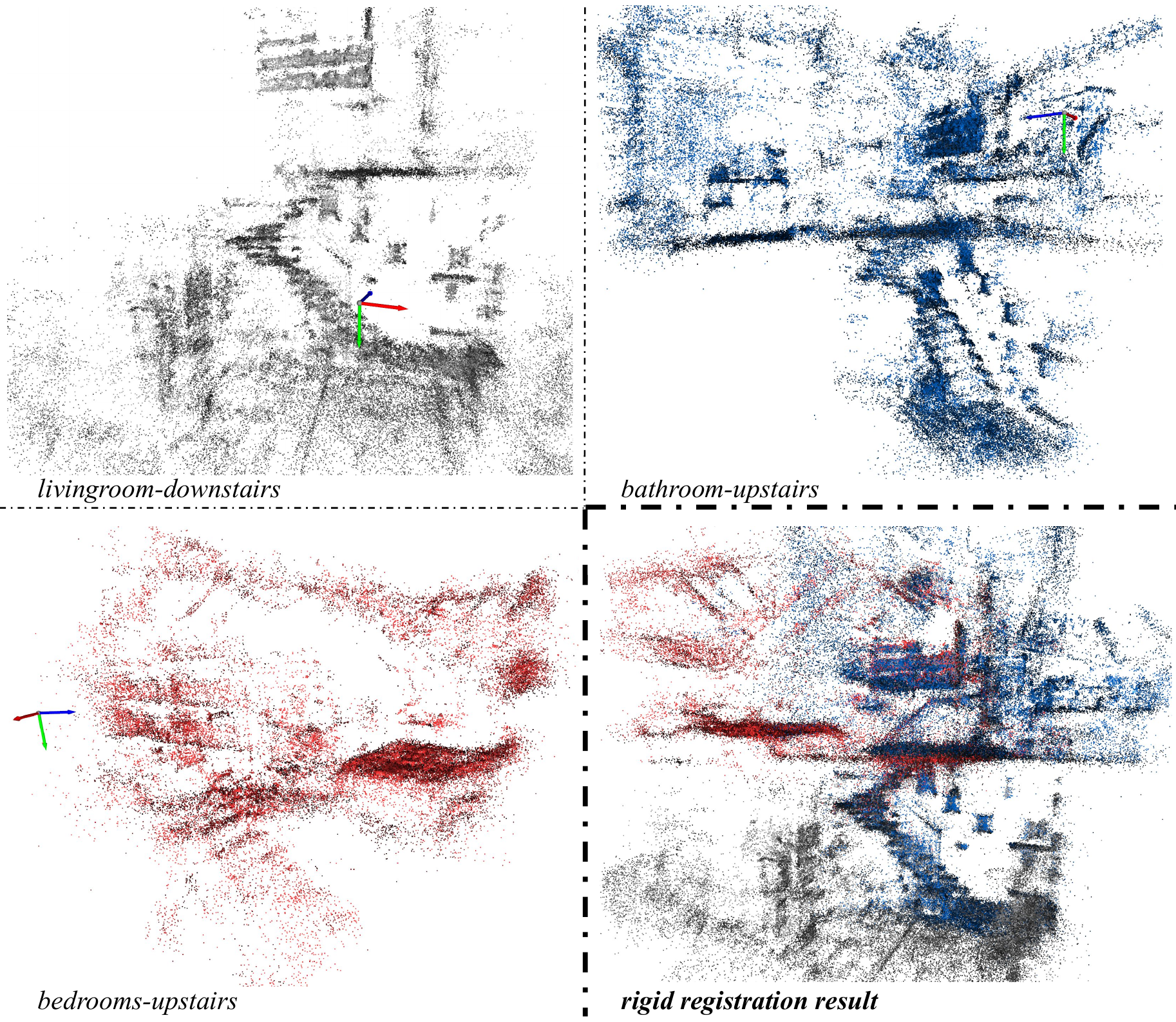}
  \vspace{-7mm}
  \caption{Estimated rigid transformations of three point clouds from our VigoHome dataset.
  DIPs can also generalise to point clouds reconstructed with the Visual-SLAM system of ARCore (Android) running on an off-the-shelf smartphone (Xiaomi Mi8).
  Three different overlapping zones of the inside of a house have been reconstructed.
  The reference frame of each point cloud, that is where the reconstruction session has started, is shown for each zone.}
  \label{fig:vigohome_qualitative}
\end{figure}

Fig.~\ref{fig:vigohome_qualitative} shows results on our dataset, i.e.~VigoHome.
In each point cloud we included the corresponding reference frame, which is where each mapping session started.
The result of a successful registration estimated using DIPs is shown in the bottom-right corner.
We can notice that the structure of the environment largely differs from that of 3DMatch and ETH datasets, and that the distribution of the points on the surfaces is much noisier that that in the 3DMatch dataset.
To quantify the registration results, we used a similar evaluation of that used in Sec.~\ref{sec:comparative_abl_study}.
For each number of sampled points we run RANSAC 100 times and compute the registration recall.
Tab.~\ref{tab:vigohome_abl} shows that with only 5K points sampled from each point cloud, 85\% of the times the three point clouds are correctly registered.
A correct registration takes about $54s$ to be processed.
We deem this a great result because it is achieved with DIPs learnt on the 3DMatch dataset.
As additional comparison, we have also quantified the registration recall using FPFH descriptors~\cite{Rusu2009}. 
However, we found that the registration fails regardless the parameters used.
So we have intentionally not included the results obtained with FPFH in the table.

\begin{table}[t]
    \tabcolsep 3pt
    \centering
    \caption{Registration recall on the VigoHome dataset as a function of the number of sampled points.}
    \label{tab:vigohome_abl}
    \begin{tabular}{lcccccc}
        \toprule
        \multirow{2}{*}{Method} & \multicolumn{5}{c}{\# sampled points} \\
        & 15000 & 10000 & 5000 & 2500 & 1000 & 500 \\
        \toprule
        DIP & .97 & .90 & .85 & .68 & .44 & .16 \\
        \bottomrule
    \end{tabular}
\end{table}

\section{Conclusions}

We presented a novel approach to learn local, compact and rotation invariant descriptors end-to-end through a PointNet-based deep neural network using canonicalised patches.
The affine transformation embedded in our network is learnt with the specific goal of improving patch canonicalisation.
We showed the importance of this step through our ablation study.
Results showed that DIPs achieve comparable performance to the state-of-the-art on the 3DMatch dataset, but that outperform the state-of-the-art by a large margin in terms of generalisation to different sensors and scenes.
We further confirmed this by capturing a new indoor dataset using the Visual-SLAM system of ARCore (Android) running on an off-the-shelf smartphone.
We observed that we can achieve good generalisation because DIPs are learnt end-to-end from the points without any hand-crafted preprocessing after canonicalisation. 
Our future research direction is to improve the canonicalisation operation \cite{Melzi2019}.

\section*{Acknowledgment}
This research has received funding from the \emph{Fondazione CARITRO - Ricerca e Sviluppo} programme 2018-2020.

\bibliographystyle{IEEEtran}
\bibliography{refs}

\end{document}